%% file: main.tex
\documentclass[runningheads]{llncs}

\usepackage{eccv}

\usepackage{eccvabbrv}

\usepackage{graphicx}
\usepackage{booktabs}
\usepackage{multirow}
\usepackage{makecell}
\usepackage{tabularx}
\usepackage[section]{placeins}  
\usepackage{pifont}
\usepackage{indentfirst}
\newcommand{\cmark}{\ding{51}}
\newcommand{\xmark}{\ding{55}}

\usepackage[accsupp]{axessibility}  

\usepackage{hyperref}

\usepackage{orcidlink}

\newif\ifappendixonly
\providecommand{\buildmode}{arxiv}
\def\appendixbuildmode{appendix}
\ifx\buildmode\appendixbuildmode\appendixonlytrue\else\appendixonlyfalse\fi

\newif\ifarxiv
\def\arxivbuildmode{arxiv}
\ifx\buildmode\arxivbuildmode\arxivtrue\else\arxivfalse\fi

\newif\ifappendixbibliography
\appendixbibliographytrue

\begin{document}

\title{SignNet-1M: Large-Scale Multilingual Sign Language Video Dataset with Downstream Benchmarks}
\titlerunning{SignNet-1M}

\author{Zhewen He\inst{1} \and Junyi Hu\inst{1} \and Haomian Huang\inst{1} \and Zhenhua Li\inst{1,2} \and Yu-Shen Liu\inst{3} \and Yi Fang\inst{1,2}}
\authorrunning{Z. He et al.}
\institute{New York University Abu Dhabi, UAE \and ChatSign Technology \and Tsinghua University, China \\
\email{zh3510@nyu.edu, yf23@nyu.edu}}

\maketitle

\begin{abstract}
Sign language models are typically trained on datasets captured under constrained conditions, with limited viewpoint, background, and signer-identity diversity, leading to poor robustness under real-world distribution shifts. We introduce \textbf{SignNet-1M}, a large-scale augmented dataset spanning ASL, CSL, and German Sign Language (DGS). SignNet-1M synthesizes realistic variations along three axes: (i) \textit{novel-view} rendering (rotation and zoom) via 3D Gaussian Splatting (3DGS), (ii) \textit{scene/ identity} editing via diffusion models for background replacement and signer substitution while preserving sign motion and linguistic content, and (iii) \textit{post-rendering} augmentations that emulate capture and compression artifacts (e.g., pose/temporal perturbations and video-level corruptions) to better match in-the-wild recordings. Beyond data release, we provide a unified benchmark suite across downstream tasks (e.g., translation and recognition) and ablations that isolate each augmentation component. Experiments across backbones show that training with SignNet-1M consistently improves generalization under cross-view, cross-background, cross-identity, and post-rendering shifts, while maintaining strong in-distribution performance. The dataset, full augmentation pipeline, and benchmark are available at \url{https://signnet.chatsign.ai/}.
\end{abstract}

\input{sections_v0305/01_introduction}
\input{sections_v0305/02_related_work}
\input{sections_v0305/04_augmentation_pipeline}
\input{sections_v0305/03_signnet_dataset}
\input{sections_v0305/05_benchmark_protocol}
\input{sections_v0305/07_experiments}
\input{sections_v0305/08_limitations_future_work}
\input{sections_v0305/09_conclusion}

\section*{Acknowledgements}
This work was supported by New York University Abu Dhabi and ChatSign Technology. The experiments were conducted using the high-performance computing resources at New York University Abu Dhabi.

\bibliographystyle{splncs04}
\bibliography{main}

\input{sections_v0305/appendix}
\end{document}

%% file: sections_v0305/01_introduction.tex
\section{Introduction}
\label{sec:intro}

Sign language translation (SLT) has advanced rapidly on standard benchmarks such as Phoenix14T~\cite{camgoz2018neural}, How2Sign~\cite{duarte2021how2sign} and CSL-Daily~\cite{zhou2021signbt}, yet these datasets are captured under constrained conditions---near-frontal cameras, studio backgrounds, and few signers---that poorly represent real-world deployment.
State-of-the-art models such as SpaMo~\cite{hwang2025spamo} and UniSign~\cite{li2025unisign} are consequently brittle to three major distribution shifts: \emph{viewpoint}, \emph{background}, and \emph{signer appearance}, as well as common image- and time-domain augmentations (e.g., photometric and compression artifacts).
For example, SpaMo trained on Phoenix14T shows a substantial BLEU-4 drop under viewpoint augmentation, highlighting a critical coverage blind spot of standard benchmarks (\cref{sec:exp_main}).

Conventional pixel-level augmentations (cropping, flipping, color jitter) cannot synthesize the structured, 3D-consistent variations that characterize viewpoint/background/identity shifts.
Recent advances in 3D Gaussian Splatting (3DGS)~\cite{kerbl20233dgs} and diffusion-based video editing~\cite{gao2025flowportal} now make it feasible to produce geometry-aware viewpoint, scene, and appearance variations at scale while preserving temporal consistency---opening the door to \emph{structure-aware} sign language data augmentation.

We introduce \textbf{SignNet-1M}, a large-scale augmented multilingual sign language dataset of ${\sim}$1M clips across three languages (ASL, DGS, CSL) drawn from seven source corpora.
Each clip is augmented along three axes:
(i)~\emph{viewpoint} via 3DGS novel-view rendering (GUAVA~\cite{zhang2025guava}),
(ii)~\emph{background} via diffusion-based scene editing (FlowPortal~\cite{gao2025flowportal} + IC-Light~\cite{zhang2025iclight}), and
(iii)~\emph{identity} via cross-reenactment signer substitution.
We additionally apply a lightweight \emph{post-rendering augmentation} stage (standard image-space transforms plus mild temporal resampling) to further broaden clip diversity.
We pair the dataset with an evaluation protocol (Orig / Zero-shot / Trained) that quantifies both the coverage blind spot of existing benchmarks and the training value of augmented data.
Our contributions are:
\begin{itemize}
    \item \textbf{SignNet-1M dataset}: ${\sim}$1M augmented clips spanning ASL, DGS, and CSL with controllable viewpoint, background, and identity variations from seven source corpora.
    \item \textbf{Scalable augmentation pipeline}: 3DGS novel-view rendering + diffusion-based scene/identity editing, plus lightweight post-rendering augmentations, engineered to preserve sign dynamics and linguistic labels.
    \item \textbf{Benchmark protocol}: a unified evaluation framework with severity-stratified robustness analysis and ablations that reveal where SLT models fail and how augmented training recovers performance.
\end{itemize}

%% file: sections_v0305/02_related_work.tex
\section{Related Work}
\label{sec:related}

\subsection{Sign language datasets and SLT pipelines.}
SLT benchmarks have expanded in scale and language coverage.
Phoenix14T~\cite{camgoz2018neural} established the standard DGS setting with aligned gloss and translation supervision, followed by ASL datasets such as How2Sign~\cite{duarte2021how2sign}, OpenASL~\cite{shi2022openasl}, and YouTube-ASL~\cite{uthus2023youtubeasl}, and CSL datasets such as CSL-Daily~\cite{zhou2021signbt} and CSL-News~\cite{li2025unisign}.
Signer-independent collections (e.g., FluentSigners-50~\cite{mukushev2022fluentsigners50}) further highlight growing interest in generalization across identities.
However, most training data remain biased toward constrained capture---near-frontal views, limited backgrounds, and restricted signer pools---making robustness to real-world shifts hard to assess in a unified manner.

SLT modeling has co-evolved with these datasets.
Early systems largely follow a gloss-supervised paradigm, translating videos via an intermediate gloss sequence~\cite{camgoz2018neural}.
Transformer-based SLT strengthened sequence modeling, e.g., Sign Language Transformers~\cite{camgoz2020sign} and STMC-Transformer~\cite{yin2020stmc}.
More recently, methods exploit stronger vision encoders and vision--language alignment: SpaMo performs gloss-free translation with prompt-based LLM decoding~\cite{hwang2025spamo}, while Uni-Sign incorporates pose/skeleton streams as explicit motion priors and a unified pre-training/fine-tuning recipe~\cite{li2025unisign}.
Yet most gains are reported on i.i.d.\ splits drawn from the same recording setups, leaving factor-specific robustness (viewpoint, background, signer identity) under-quantified.
SignNet-1M complements this line with controlled, factorized visual shifts and a unified protocol for diagnosing robustness.

\subsection{Data augmentation and robustness under distribution shifts.}
Data augmentation is central to robustness in visual recognition, including policy-based transforms (RandAugment~\cite{cubuk2020randaugment}), composition (AugMix~\cite{hendrycks2020augmix}), and sample mixing (MixUp~\cite{zhang2018mixup}, CutMix~\cite{yun2019cutmix}).
For video, temporal coherence further constrains valid augmentations, motivating video-specific invariances and actor/scene mixing (e.g., ActorCutMix~\cite{zou2021videoaug}).
However, such pixel/patch-level transforms are insufficient for SLT: sign meaning depends on subtle hand/face motion that can be corrupted by aggressive appearance edits, and real deployments exhibit structured shifts (viewpoint, background, unseen signer) that are difficult to synthesize with frame-wise operations alone.

Sign-language augmentation has mainly targeted data scarcity or alignment rather than factor-controlled shifts.
Sign Back-Translation (SignBT)~\cite{zhou2021signbt} leverages monolingual text to enrich SLT pairs, and other cross-modality strategies introduce auxiliary cues, but they do not provide explicit, controllable variation along viewpoint/background/identity axes at scale.
In contrast, SignNet-1M generates factorized, controllable visual variations while preserving motion and linguistic annotations, enabling severity-stratified robustness re-training and evaluation.

\subsection{3D-aware rendering and diffusion-based video editing.}
Controllable generation enables structured variations beyond conventional augmentation.
3D Gaussian Splatting (3DGS)~\cite{kerbl20233dgs} supports efficient novel-view synthesis with explicit camera control.
GUAVA~\cite{zhang2025guava} builds animatable upper-body avatars via SMPL-X~\cite{pavlakos2019smplx} and FLAME~\cite{li2017flame}, enabling geometry-consistent viewpoint manipulation and cross-identity reenactment through disentangled identity and motion parameters.

Diffusion-based video generation/editing has also progressed rapidly, including Tune-A-Video~\cite{wu2023tuneavideo}, Text2Video-Zero~\cite{khachatryan2023text2videozero}, and Video-P2P~\cite{liu2024videop2p}.
For practical scene editing, FlowPortal~\cite{gao2025flowportal} performs temporally consistent relighting/background replacement via residual-corrected flow in latent space, while IC-Light~\cite{zhang2025iclight} provides illumination-aware conditioning.
SignNet-1M assembles these components into a sign-language-tailored augmentation pipeline: 3D-aware avatars provide controllable viewpoint/identity variation, and diffusion editing introduces diverse, temporally consistent background changes---together yielding factorized shifts that can be systematically benchmarked.

%% file: sections_v0305/04_augmentation_pipeline.tex

\section{Methodology}
\label{sec:augmentation}

\begin{figure}[t]
    \centering
    \includegraphics[width=1\linewidth, height=0.55\textheight, keepaspectratio]{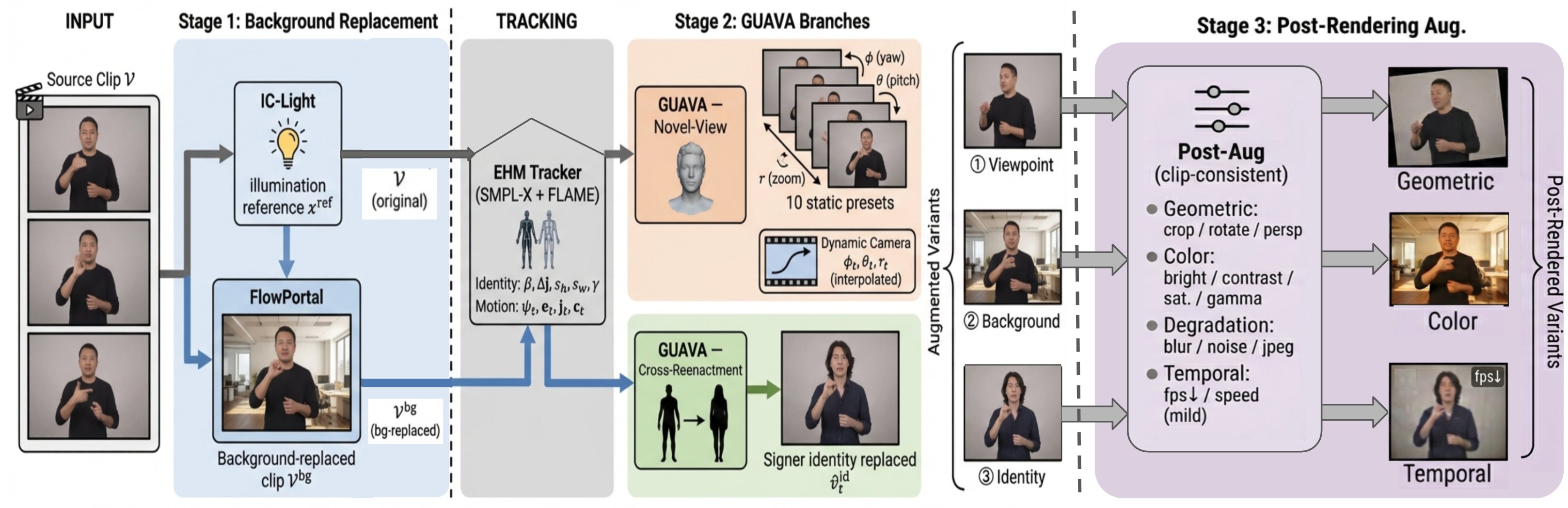}
    \caption{Overview of our augmentation framework. Stage~1: background replacement (FlowPortal~\cite{gao2025flowportal} + IC-Light~\cite{zhang2025iclight}). Stage~2: EHM-Tracker~\cite{zhang2025ehmtracker} tracks each stream to SMPL-X/FLAME parameters, then GUAVA~\cite{zhang2025guava} performs novel-view rendering (static/dynamic cameras) and cross-identity reenactment. Stage~3: post-rendering augmentations (video-consistent image-space transforms and mild temporal resampling).}
    \label{fig:pipeline_overview}
\end{figure}

\begin{figure}[t]
    \centering
    \includegraphics[height=0.46\textheight, keepaspectratio]{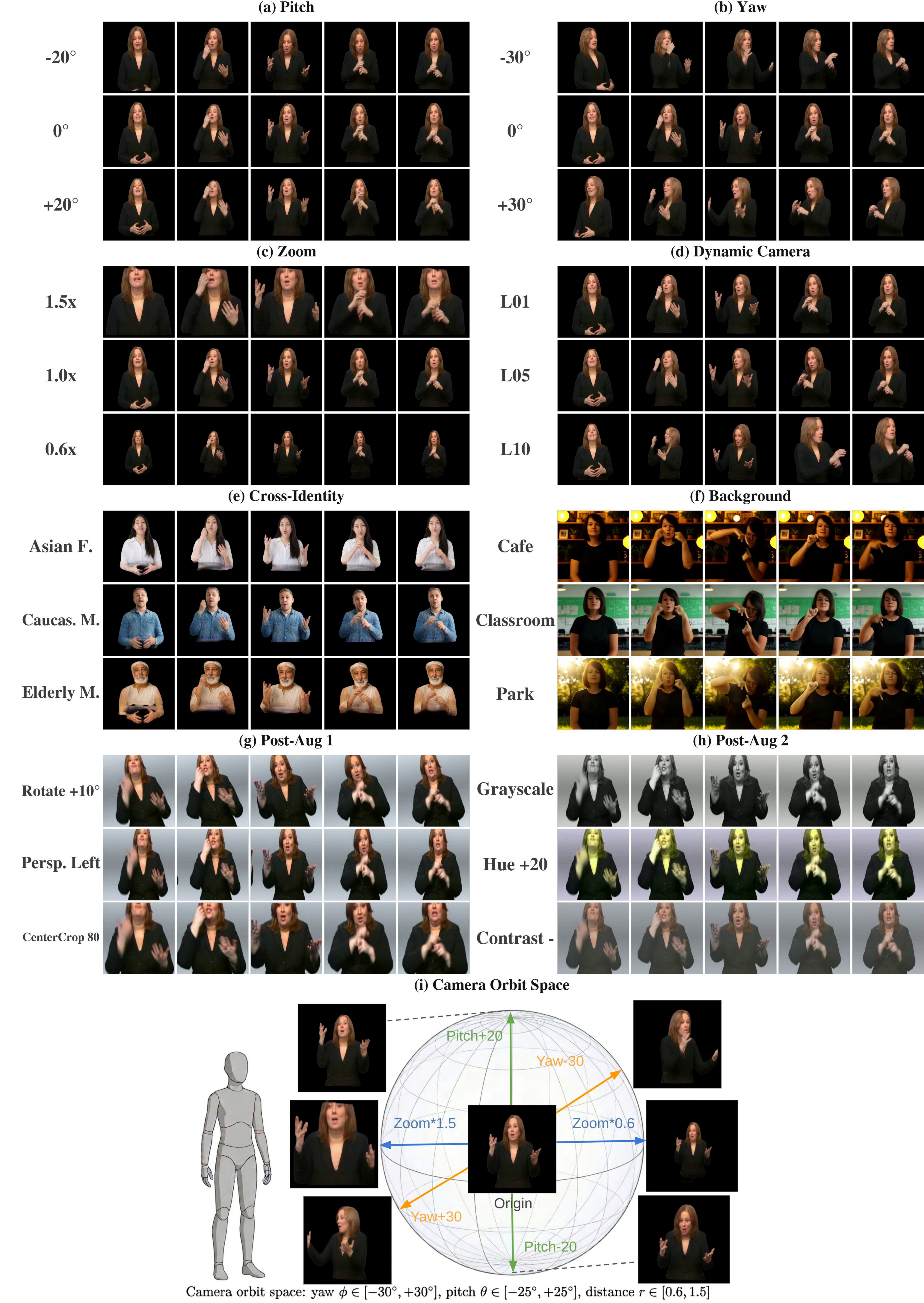}
    \caption{\small\textbf{Qualitative augmentation examples.} (a--d) Novel-view camera augmentations (pitch, yaw, zoom, dynamic). (e) Cross-identity reenactment. (f) Background replacement. (g--h) Post-rendering augmentations (geometric/photometric/temporal). (i) Camera orbit space visualization.}
    \label{fig:augmentation_examples}
\end{figure}

We construct SignNet-1M by applying controlled factor shifts to each source clip along three axes--- \emph{viewpoint}, \emph{background}, and \emph{signer identity}---followed by lightweight \emph{post-rendering} augmentations.
For notational convenience, we denote a clip as $\mathcal{V}=\{v_t\}_{t=1}^{T}$ when needed.
Fig.~\ref{fig:pipeline_overview} summarizes the procedure, and Fig.~\ref{fig:augmentation_examples} illustrates typical outcomes (a--h).

Concretely, we optionally replace the background to create an alternative stream $\mathcal{V}^{\text{bg}}$ (Stage~1), track the signer to a parametric representation and render novel views and cross-identity reenactment on both $\mathcal{V}$ and $\mathcal{V}^{\text{bg}}$ (Stage~2), and finally apply lightweight post-rendering augmentations that preserve linguistic annotations (Stage~3), together spanning all three factor axes (Fig.~\ref{fig:augmentation_examples}a--h).

\subsection{Novel-View Rendering via 3DGS}
\label{sec:aug_3dgs}

\textbf{Parametric tracking.}
To enable 3D-aware view synthesis while preserving the original signing motion, we first track each clip to a compact parametric representation.
EHM-Tracker~\cite{zhang2025ehmtracker} fits SMPL-X~\cite{pavlakos2019smplx} + FLAME~\cite{li2017flame} per frame and factorizes clip-level identity from time-varying motion:

\begin{equation}
\label{eq:tracking}
\theta_t=\bigl(\underbrace{\beta,\Delta\mathbf{j},s_h,s_w,\gamma}_{\text{identity}},
\underbrace{\psi_t,\mathbf{e}_t,\mathbf{j}_t,\mathbf{c}^{\text{src}}_t}_{\text{motion}}\bigr),
\end{equation}
where identity terms are constant within a clip and motion terms vary over time.

\textbf{3D-aware rendering and compositing.}
Given the tracked pose $\psi_t$ as input, GUAVA~\cite{zhang2025guava} renders the avatar under a camera preset $\mathbf{c}^{(k)}$ and alpha-composites it with the current background stream:
\begin{equation}
\label{eq:composite}
\hat{v}_t^{(k)}=
\mathcal{R}(\widetilde{\mathcal{G}}_t,\mathbf{c}^{(k)})\odot \alpha_{t}^{\text{rend},(k)}
+ b_t\odot\bigl(1-\alpha_{t}^{\text{rend},(k)}\bigr),
\end{equation}
where $\alpha_{t}^{\text{rend},(k)}$ is the rendered alpha matte and $b_t$ is the background frame from the current stream (either $v_t$ or $v_t^{\text{bg}}$).
This produces the novel-view variants in Fig.~\ref{fig:augmentation_examples}a--d.

\textbf{Viewpoint severity presets.}
We define $K{=}10$ spherical-camera presets $(\phi,\theta,r)$ with LookAt $=[0,0.75,0]$, indexed by increasing perturbation magnitude (Tab.~\ref{tab:difficulty_levels}).
For each level $\ell\in\{1,\dots,10\}$, we sample $(\phi,\theta,r)$ uniformly within the corresponding ranges.
We additionally generate a dynamic-camera trajectory by linearly interpolating between two endpoints sampled from the same level (Fig.~\ref{fig:augmentation_examples}d), enabling temporally varying viewpoint shifts. The camera orbit space visualization is also shown in Fig.~\ref{fig:augmentation_examples}i.

\subsection{Background Replacement}
\label{sec:aug_background}

We estimate a temporally consistent soft alpha $\alpha_t^{\text{bg}}$ via video matting, and apply illumination-aware background replacement using IC-Light~\cite{zhang2025iclight} and FlowPortal~\cite{gao2025flowportal}.
Edits are blended using $\alpha_t^{\text{bg}}$ to preserve the signer’s motion and geometry while altering background and lighting (Fig.~\ref{fig:augmentation_examples}f).

\noindent\textbf{Lighting severity.}
Since illumination changes can affect sign-language perception, we stratify background-edited clips into 10 severity levels (L1--L10) using a signer-region photometric shift score (original vs.\ edited).
The exact score definition is provided in the appendix, and the bin ranges are summarized in Tab.~\ref{tab:difficulty_levels}.

\subsection{Cross-Identity Reenactment}
\label{sec:aug_identity}

Using the factorization in \cref{eq:tracking}, we perform cross-identity reenactment by combining identity from a source image with motion from a target clip.
Specifically, we take $\Theta^{\text{id}}$ from the source and $\Theta_t^{\text{mot}}$ from the target to form $\tilde{\theta}_t=(\Theta^{\text{id}},\Theta_t^{\text{mot}})$, then render with the target camera/background stream using the same compositor as in \cref{eq:composite}.
This yields the cross-identity variants shown in Fig.~\ref{fig:augmentation_examples}e.

\subsection{Post-Rendering Augmentations}
\label{sec:aug_post}

Finally, we apply lightweight \emph{post-rendering} augmentations to rendered/edited clips to broaden appearance diversity while preserving linguistic annotations (Fig.~\ref{fig:augmentation_examples}g--h).
We use video-consistent image-space transforms (e.g., crop/rotate, color/contrast jitter, mild blur/compression) and mild temporal resampling (e.g., small speed jitter or frame dropping), which are inexpensive yet effective for increasing low-level variability without changing the underlying sign content.

%% file: sections_v0305/03_signnet_dataset.tex

\section{SignNet-1M Dataset}
\label{sec:dataset}

\subsection{Basic Information}
\label{sec:dataset_sources}

\textbf{Data Source.} SignNet-1M comprises $\sim$1M \textbf{augmented} sign-language clips spanning three languages: DGS, ASL, and CSL, built from seven public or self-collected corpora to diversify lexicon and recording conditions.

\begin{figure}[t]
    \centering
    \includegraphics[width=0.95\linewidth]{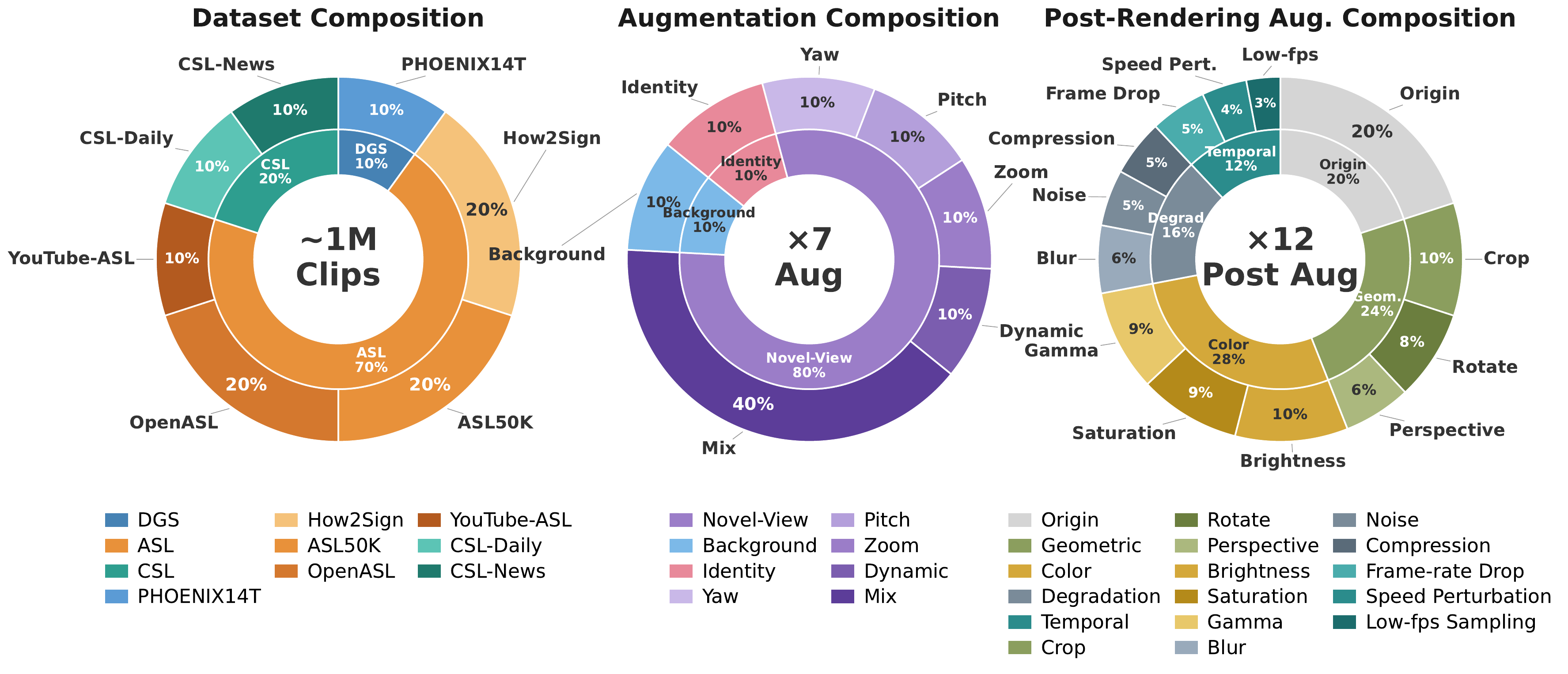}
    \caption{\textbf{SignNet-1M composition.} \textbf{Left:} source-language breakdown (ASL 70\%, CSL 20\%, DGS 10\%). \textbf{Middle:} generative augmentation mix ($\times$7): novel-view (80\%; yaw/pitch/zoom/dynamic/mix) plus background (10\%) and identity (10\%) editing. \textbf{Right:} post-rendering augmentation mix ($\times$12): origin (20\%) plus geometric/color/degradation/temporal augmentations.}
    \label{fig:dataset_composition}
\end{figure}

The DGS subset ($\sim$100K) uses Phoenix14T~\cite{camgoz2018neural}.
The ASL subset ($\sim$700K) combines How2Sign~\cite{duarte2021how2sign}, OpenASL~\cite{shi2022openasl}, ASL50K~\cite{asl50k} (commercially licensed; our augmented clips in SignNet-1M can be distributed for non-commercial research use), and YouTube-ASL~\cite{uthus2023youtubeasl}.
The CSL subset ($\sim$200K) includes CSL-Daily~\cite{zhou2021signbt} and CSL-News~\cite{li2025unisign}.
\cref{fig:dataset_composition} summarizes source and augmentation breakdowns.

\textbf{Augmentation composition.}
We apply a generative augmentation \emph{framework} to each source clip: (i) 3DGS novel-view synthesis (80\%) with yaw/pitch/zoom/ dynamic/mixed augmentations (10/10/10/10/40\%), and (ii) diffusion-based appearance editing (20\%) split into background (10\%) and identity (10\%).
We further apply lightweight \emph{post-rendering} augmentations (video-consistent image-space transforms and mild temporal resampling) to broaden diversity without changing linguistic annotations.
Details are in \cref{sec:augmentation}.
Per-language counts and the per-clip computational cost of each augmentation stage are reported in Appendix~A.

\textbf{Data access and license.}
We publicly release SignNet-1M together with the \emph{full augmentation pipeline} (source code and per-stage configuration files) at \url{https://signnet.chatsign.ai/}, including (i) train/val/test splits (8:1:1, split at the source-clip level to prevent leakage), (ii) augmentation metadata (factor axes and severity levels), and (iii) evaluation scripts to reproduce the Orig/Zero-shot/Trained protocol.
Releasing the per-stage configs and code lets practitioners reproduce and extend every augmentation axis.
We will distribute the data under a non-commercial license and provide a standardized data-card describing sources, intended use, and known limitations.

\subsection{Preprocessing and Annotation}
\label{sec:dataset_preprocess}

\textbf{Preprocessing.}
For YouTube-ASL and OpenASL, we segment clips using sentence-level timestamps \cite{uthus2023youtubeasl, shi2022openasl}.
We apply confidence-based pose QC with EHM-Tracker and DWPose \cite{zhang2025ehmtracker, yang2023dwpose}: keypoints with confidence $<0.3$ are treated as invisible; frames are discarded if either hand has mean confidence $<0.7$ or if severe self-occlusion is detected (two-hand distance MSE $<3072$).

\textbf{Annotation.}
All linguistic annotations (gloss/translation/timestamps) are inherited from the source corpora unchanged.

\subsection{Dataset Statistics}
\label{sec:dataset_stats}

\cref{tab:dataset_comparison} compares SignNet-1M with prior translation datasets. The count-weighted average clip length is 7.41\,s; per-dataset statistics are provided in the appendix.

\begin{table*}[t]
    \centering
    \caption{Comparison with existing sign language translation datasets.
    \emph{Aug.}: synthetic augmentation; \emph{Ann.}: G\,=\,gloss, T\,=\,translation.}
    \label{tab:dataset_comparison}
    \small
    \begin{tabular}{llrrrcc}
        \toprule
        Dataset & Language & Clips & Duration (h) & Signers & Aug. & Ann. \\
        \midrule
        Phoenix14T~\cite{camgoz2018neural}   & DGS & 8.26K & 11 & 9   & \xmark & G+T \\
        How2Sign~\cite{duarte2021how2sign}   & ASL & 35.00K & 80 & 11  & \xmark & T \\
        OpenASL~\cite{shi2022openasl}        & ASL & 98.42K & 288 & $\sim$220 & \xmark & T \\
        YouTube-ASL~\cite{uthus2023youtubeasl} & ASL & 610.19K & 984 & $\sim$2.5K & \xmark & T \\
        CSL-Daily~\cite{zhou2021signbt}      & CSL & 20.65K & 23 & $\sim$10  & \xmark & G+T \\
        CSL-News~\cite{li2025unisign}        & CSL & 751.32K & 1,985 & $\sim$20 & \xmark & T \\
        \midrule
        SignNet-1M (ours) & DGS+ASL+CSL & \textbf{1M} & \textbf{2,058} & \textbf{$\sim$10K}\footnotemark & \cmark & G+T \\
        \bottomrule
    \end{tabular}
\end{table*}

\footnotetext{The reported number of signers in SignNet-1M includes \emph{synthetic identities} introduced by cross-identity reenactment, in addition to original signers from the source corpora.}

%% file: sections_v0305/05_benchmark_protocol.tex
\section{Benchmark Suite and Augmented Evaluation Protocol}
\label{sec:benchmark}


\subsection{Downstream Tasks}
\label{sec:benchmark_tasks}

We evaluate on two complementary sign language understanding tasks.

\textbf{Sign Language Translation (SLT)} maps a video clip to a spoken-language sentence and is the primary task of our benchmark.
We report BLEU-4~\cite{papineni2002bleu} ($\uparrow$) as the main metric throughout the paper; ROUGE-L~\cite{lin2004rouge} and ChrF~\cite{popovic2015chrf} results are provided in the supplementary material. All included source corpora provide sentence-level translations, so SLT experiments cover the full SignNet-1M benchmark across DGS, ASL, and CSL.

\textbf{Sign Language Recognition (SLR)} maps a video clip to a gloss sequence and is evaluated by Word Error Rate (WER\%, $\downarrow$).
Because gloss annotations are only available in Phoenix14T~\cite{camgoz2018neural} and CSL-Daily~\cite{zhou2021signbt}, SLR experiments are restricted to the DGS and CSL sub-datasets of SignNet-1M.

\subsection{Evaluation Protocol}
\label{sec:benchmark_aug_eval}

Each task is evaluated under four settings to measure both robustness gaps in existing data and the training value of SignNet-1M:
\begin{itemize}
    \item \textbf{Setting A (Orig.)}: official split and training setup on the source benchmark (train/test on the source benchmark).
    \item \textbf{Setting B (Zero-shot)}: evaluate the Orig.-trained model on the SignNet-1M test split.
    \item \textbf{Setting C (Trained)}: train on the SignNet-1M trainset and evaluate on the SignNet-1M testset.
    \item \textbf{Setting D}: train on SignNet-1M and evaluate on the \emph{original} benchmark test set, measuring whether augmented training helps under the natural source distribution.
\end{itemize}
The SignNet-1M test split is obtained by augmenting the \emph{source test set}; splits are made at the source-clip level, so no train/test leakage occurs between Settings.
We define \textbf{Gap}$=\text{B}-\text{A}$ (Zero-shot$-$Orig.) and \textbf{Gain}$=\text{C}-\text{B}$ (Trained$-$Zero-shot) (higher is better); for lower-is-better metrics (e.g., WER\%), we flip the sign.
Negative Gap indicates sensitivity to viewpoint/background/identity/post-rendering shifts, while positive Gain shows that SignNet-1M improves robustness under such shifts.

\subsection{Severity-Stratified Evaluation Factors}
\label{sec:benchmark_factors}

Beyond aggregate performance, we stratify the SignNet-1M test set by augmentation severity to reveal how robustness degrades as visual conditions become more challenging.
Severity levels are defined along a viewpoint axis (static \& dynamic) and a scene-lighting axis, and summarised in
\cref{tab:difficulty_levels}.

\begin{table}[h]
    \centering
    \caption{Ranges of parameters and severity-level definitions for stratified evaluation.}
    \label{tab:difficulty_levels}
    \small
    \setlength{\tabcolsep}{4pt}
    \renewcommand{\arraystretch}{1.1}
    \begin{tabular}{@{}llp{6.6cm}@{}}
        \toprule
        Axis & Levels & Definition \\
        \midrule
        Viewpoint (static)
        & L1--L10
        & $|\phi|: 0^{\circ}\!\to\!30^{\circ},\;
           |\theta|: 0^{\circ}\!\to\!25^{\circ},\;
           r: [1.0,1.0]\!\to\![0.6,1.5]$ \\

        Viewpoint (dynamic)
        & L1--L10
        & Sample two endpoints from the corresponding static L$k$ range and linearly interpolate $(\phi_t,\theta_t,r_t)$ over time. \\

        Lighting
        & L1--L10
        & $\hat{s}_{\text{light}}\in[0,1]$: normalized signer-region photometric shift, binned into 10 equal-width levels. \\
        \bottomrule
    \end{tabular}
\end{table}

\textbf{Viewpoint severity.}
We evaluate on each of the $K{=}10$ camera presets defined in \cref{sec:aug_3dgs}, indexed by increasing augmentation magnitude, plus the dynamic-camera setting, yielding 11 points in total.

\textbf{Scene severity.}
For diffusion-based background editing, we define 10 linearly spaced levels along a single axis of lighting variation.
Lighting levels are computed from a normalized signer-region photometric severity score, with parameter ranges in Tab.~\ref{tab:difficulty_levels}.

Both Zero-shot and Trained results are reported per level in \cref{sec:exp_difficulty}, enabling fine-grained diagnosis of where and how model performance degrades.

%% file: sections_v0305/07_experiments.tex
\section{Experiments}
\label{sec:experiments}

\subsection{Experimental Setup}
\label{sec:exp_setup}

\textbf{Datasets.}
We evaluate on four SignNet-1M sub-datasets spanning DGS, ASL, and CSL: Phoenix14T, How2Sign, OpenASL, and CSL-Daily.
All results follow the unified three-condition protocol in \cref{sec:benchmark_aug_eval}.

\textbf{Models.}
We benchmark representative SOTA backbones from distinct model families.
For translation, we include SpaMo~\cite{hwang2025spamo} and additionally report GASLT~\cite{yin2023gaslt} and FPT4SLT~\cite{wong2024sign2gpt} where applicable.
For unified recognition/translation, we evaluate UniSign~\cite{li2025unisign}.
For recognition on Phoenix14T, we further include an online CSLR baseline~\cite{zuo2024towards}.

\textbf{Training and evaluation.}
For fair comparison, we follow each method's official training and evaluation setup and vary only the data (Orig.\ vs.\ SignNet-1M).
We report BLEU-4 for translation and WER\% for recognition; additional metrics and all method-specific hyperparameters are reported in the Appendix.
We further conduct severity-stratified evaluations (\cref{sec:exp_difficulty}), ablations (\cref{sec:exp_ablations}), scaling experiments (\cref{sec:scaling}), and visual-quality checks (\cref{sec:exp_quality}).

\begin{table}[t]
    \centering
    \small
    \caption{Sign language translation metric BLEU-4 ($\uparrow$) on sub-datasets of SignNet-1M.
    Evaluation conditions and $\Delta$ metrics are defined in \cref{sec:benchmark_aug_eval}.
    Setting D trains on SignNet-1M and evaluates on the original (Orig.) test set; A/B/C/D are defined in \cref{sec:benchmark_aug_eval}.}
    \label{tab:main_exp}
    \small
    \setlength{\tabcolsep}{4pt}
    \resizebox{\textwidth}{!}{%
    \begin{tabular}{ll c cc c cc}
        \toprule
        & & Source & \multicolumn{2}{c}{SignNet-1M} & \multicolumn{1}{c}{Orig.\,test} & \multicolumn{2}{c}{$\Delta$} \\
        \cmidrule(lr){3-3} \cmidrule(lr){4-5} \cmidrule(lr){6-6} \cmidrule(lr){7-8}
        Sub-dataset & Method & A (Orig.) & B (Zero-shot) & C (Trained) & D (Trained$\to$Orig.) & Gap & Gain \\
        \midrule
        \multirow{4}{*}{Phoenix14T}
          & SpaMo~\cite{hwang2025spamo}  & 22.49 & 7.81  & 18.98 & 27.78 & -14.68 & +11.17 \\
          & GASLT~\cite{yin2023gaslt}    & 15.74  & 1.26  & 14.95  & -- & -14.48   & +13.69 \\
          & FPT4SLT~\cite{wong2024sign2gpt}  & 19.18  & 7.25  & 18.72  & -- & -11.93  & +11.47 \\
        \midrule
        \multirow{2}{*}{How2Sign}
          & SpaMo~\cite{hwang2025spamo}  & 10.11 & 6.25  & 10.36 & 18.48 & -3.86  & +4.11 \\
          & UniSign~\cite{li2025unisign} & 14.50 & 6.47  & 12.83  & -- &  -8.03  & +6.36 \\
        \midrule
        \multirow{2}{*}{OpenASL}
         & SpaMo~\cite{hwang2025spamo} & 10.32 & 0.34 & 9.71 & -- & -9.98 & +9.37 \\
          & UniSign~\cite{li2025unisign} & 22.67 & 8.12  & 22.83  & -- &  -14.55  & +14.71 \\
        \midrule
        \multirow{2}{*}{CSL-Daily}
          & SpaMo~\cite{hwang2025spamo}  & 20.55 & 12.70 & 21.32 & 20.93 & -7.85  & +8.62 \\
          & UniSign~\cite{li2025unisign} & 25.27 & 13.47 & 21.66 & -- & -11.80 & +8.19 \\
          & GASLT~\cite{yin2023gaslt}    &  4.07  &  0.12  & 3.68   & -- &  -3.95  & +3.56 \\
        \bottomrule
    \end{tabular}%
    }
\end{table}

\begin{table}[t]
    \centering
    \small
    \caption{Sign language recognition metric WER\% ($\downarrow$) on sub-datasets of SignNet-1M.
    Evaluation conditions and $\Delta$ metrics are defined in \cref{sec:benchmark_aug_eval}.}
    \label{tab:main_exp_slr}
    \small
    \begin{tabular}{ll c cc cc}
        \toprule
        & & Source & \multicolumn{2}{c}{SignNet-1M} & \multicolumn{2}{c}{$\Delta$} \\
        \cmidrule(lr){3-3} \cmidrule(lr){4-5} \cmidrule(lr){6-7}
        Sub-dataset & Method & Orig. & Zero-shot & Trained & Gap & Gain \\
        \midrule
        Phoenix14T
          & Online-CSLR~\cite{zuo2024towards} & 22.21\% & 51.45\% & 28.19\% & -29.24\% & +23.26\% \\
        \midrule
        CSL-Daily
          & UniSign~\cite{li2025unisign} & 28.20\% & 56.90\% & 30.74\% & -28.70\% & +26.16\% \\
        \bottomrule
    \end{tabular}
\end{table}

\subsection{Translation \& Recognition Performance}
\label{sec:exp_main}
We evaluate SignNet-1M on sign language translation (SLT) and recognition (SLR) under the unified protocol in \cref{sec:benchmark_aug_eval}.

\textbf{Translation performance.}
\cref{tab:main_exp} reports BLEU-4 under \textit{Orig.}, \textit{Zero-shot}, and \textit{Trained} settings.
Zero-shot BLEU-4 drops sharply across datasets even for strong baselines (e.g., Phoenix14T: $-11.93$ to $-14.68$; OpenASL: $-9.98$ to $-14.55$), indicating sensitivity to the factor shifts in SignNet-1M.
Training on SignNet-1M consistently recovers performance with large gains across languages and architectures (e.g., Phoenix14T: $+11.17$ to $+13.69$; OpenASL: $+9.37$ to $+14.71$), demonstrating improved robustness under challenging visual conditions.
Crucially, models trained on SignNet-1M also improve on the \emph{original} benchmark test sets (Setting D vs.\ Setting A): SpaMo gains $+5.29$, $+8.37$, and $+0.38$ BLEU-4 on Phoenix14T, How2Sign, and CSL-Daily, showing the augmentation helps under natural distributions, not only constructed shifts.
Across three typical model-family specifications (RGB gloss-free, RGB+Pose gloss-based SLT, and pose-based recognition), training on SignNet-1M consistently improves over original-dataset training (summarized in the supplementary material), indicating the gains are not tailored to a single architecture.

\textbf{Recognition performance.}
\cref{tab:main_exp_slr} shows a similar trend for WER\% ($\downarrow$).
Zero-shot recognition degrades substantially on SignNet-1M (Phoenix14T: $22.21\%\!\rightarrow\!51.45\%$; CSL-Daily: $28.20\%\!\rightarrow\!56.90\%$), while training on SignNet-1M markedly reduces WER (to $28.19\%$ and $30.74\%$, respectively), yielding large gains ($+23.26\%$ and $+26.16\%$).
This confirms that SignNet-1M improves robustness for both translation and recognition, rather than only boosting average-case accuracy.

\begin{figure}[t]
    \centering
    \includegraphics[height=0.30\textheight, keepaspectratio]{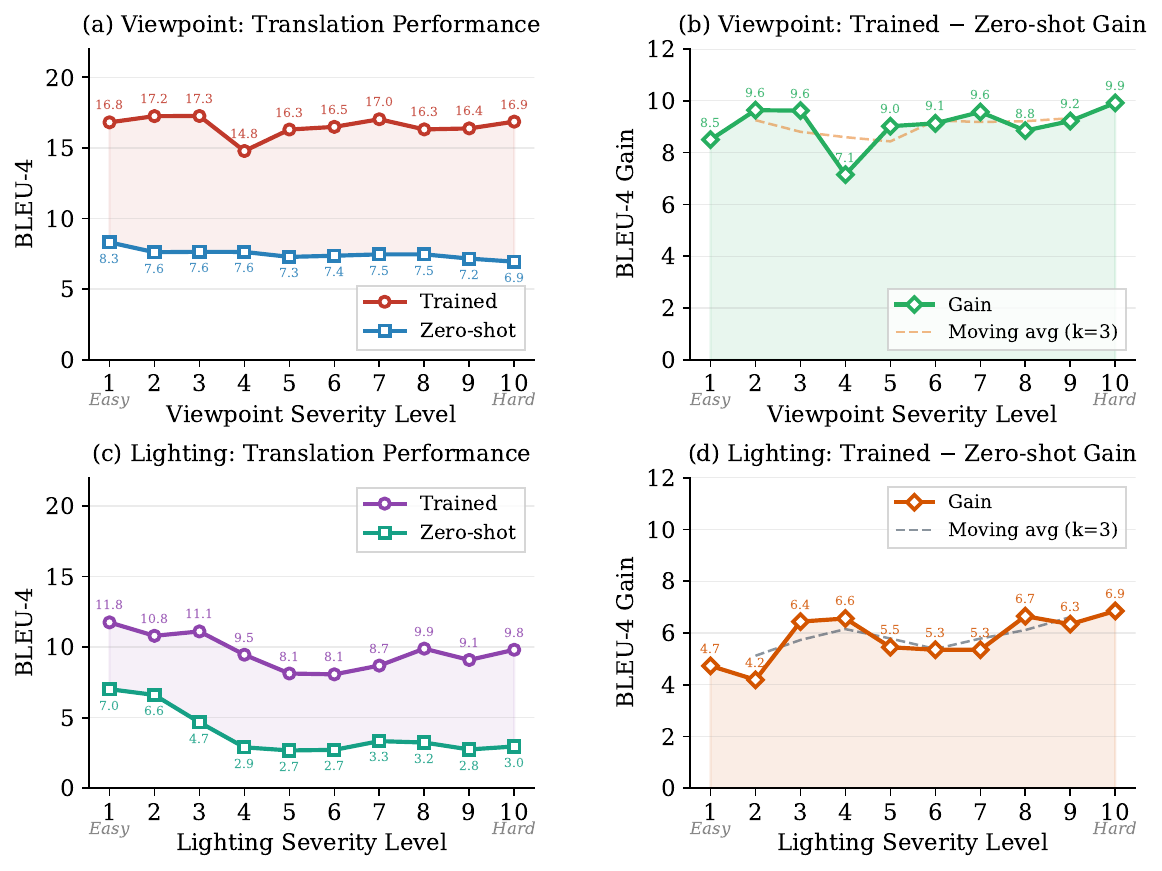}
    \caption{Stratified SLT performance across severity levels on Phoenix14T. (a,c) BLEU-4 of the zero-shot and the trained model. (b,d) Trained$-$zero-shot gain in BLEU-4. Viewpoint levels (L1--L10) follow increasing yaw/pitch/zoom augmentations, and lighting levels (L1--L10) are defined by a normalized photometric shift defined in Tab.~\ref{tab:difficulty_levels}.}
    \label{fig:difficulty}
\end{figure}

\subsection{Experiments Under Increasing Severity}
\label{sec:exp_difficulty}
Following the severity stratification in \cref{sec:benchmark_factors}, we report BLEU-4 across severity levels (L1--L10) under \textit{zero-shot} and \textit{trained} settings on SignNet-1M (Phoenix14T sub-dataset, SpaMo) \cite{hwang2025spamo}.
\cref{fig:difficulty} summarizes robustness under increasing viewpoint augmentations and lighting shifts.

\textbf{Viewpoint.}
As viewpoint severity increases (\cref{fig:difficulty}a), zero-shot BLEU-4 steadily degrades (8.31 $\rightarrow$ 6.94), while training yields higher scores with a much flatter drop.
Accordingly, the training gain grows at harder levels (\cref{fig:difficulty}b; 8.50 $\rightarrow$ 9.92), indicating improved robustness under severe viewpoint shifts.

\textbf{Lighting.}
Lighting shifts induce a larger zero-shot collapse (\cref{fig:difficulty}c; 7.02 $\rightarrow$ 2.96), whereas training substantially mitigates the drop and remains consistently stronger across levels.
The gain also increases with severity (\cref{fig:difficulty}d; 4.73 $\rightarrow$ 6.85), suggesting training is particularly effective under extreme illumination changes. Overall, larger gains at harder levels show that covering such shifts is essential for in-the-wild robustness, motivating SignNet-1M.

\begin{table}[t]
    \centering
    \small
    \setlength{\tabcolsep}{3pt}      
    \renewcommand{\arraystretch}{0.92} 
    \caption{Ablation of augmentation components (BLEU-4).
    Results are reported on sub-datasets of SignNet-1M (Phoenix14T and How2Sign) under zero-shot and trained protocols using SpaMo method \cite{hwang2025spamo}.}
    \label{tab:ablation}
    \small
    \newcommand{\abldash}{%
        \multicolumn{5}{@{}l@{}}{%
            \leavevmode\leaders\hbox{\rule[0.5ex]{0.7em}{0.2pt}\hspace{0.45em}}\hfill\kern0pt%
        } \\
    }
    \begin{tabular}{lcccc}
        \toprule
        & \multicolumn{2}{c}{Phoenix14T} & \multicolumn{2}{c}{How2Sign} \\
        \cmidrule(lr){2-3} \cmidrule(lr){4-5}
        Configuration & Zero-shot & Trained & Zero-shot & Trained \\
        \midrule
        \textit{Novel View}         & 7.10 & 18.85 & 6.09 & 10.32 \\
        \quad + yaw                 & 8.17 & 18.59 & 7.09 & 10.78 \\
        \quad + pitch               & 6.98 & 18.36 & 4.06 &  9.04 \\
        \quad + zoom                & 8.89 & 18.98 & 7.81 & 11.05 \\
        \quad + dynamic camera      & 6.87 & 17.13 & 6.16 &  8.70 \\
        \abldash
        \textit{Scene editing}      & 4.91 & 12.53 & 3.28 &  9.26 \\
        \abldash
        \textit{Identity editing}   & 13.06 & 22.04 & 8.67 & 12.54 \\
        \abldash
        \textit{Post-rendering aug.} & 5.37 & 18.16 & 2.51 & 10.17 \\
        \abldash
        \textit{Full SignNet-1M}    & 7.81 & 18.98 & 6.25 & 10.36 \\
        \bottomrule
    \end{tabular}
\end{table}

\subsection{Ablation Study}
\label{sec:exp_ablations}
\cref{tab:ablation} analyzes the contribution of individual augmentation components in SignNet-1M on Phoenix14T and How2Sign under both zero-shot and trained protocols.
Among the novel-view factors, \texttt{zoom} yields the most consistent gains, while \texttt{dynamic camera} is less stable, indicating a harder shift that benefits from careful severity control.
For appearance variation, \textit{identity} editing consistently outperforms \textit{scene} editing alone (e.g., $13.06/22.04$ vs.\ $4.91/12.53$ on Phoenix14T), highlighting signer diversity as a more important robustness source than background change.
\textit{Post-rendering augmentation} adds a complementary gain on top of rendering-based augmentation.
Overall, the gains of SignNet-1M arise from complementary robustness factors rather than any single augmentation component.

\begin{figure}[t]
    \centering
    \includegraphics[height=0.21\textheight, keepaspectratio]{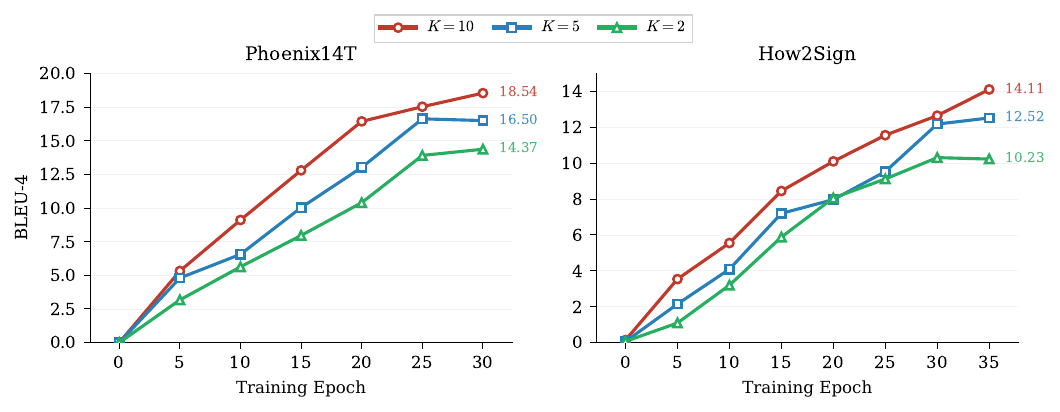}
    \caption{\textbf{Scaling with augmentation factor $K$.} BLEU-4 vs. training epoch when training with increasing augmentation scale factors ($K{=}2,5,10$). Results are shown on Phoenix14T (left) and How2Sign (right).}
    \label{fig:scaling}
\end{figure}

\subsection{Scaling Experiments}
\label{sec:scaling}
We study how translation performance scales with the amount of augmented data by training with increasing scale factors $K$ ($K{=}2,5,10$; \cref{fig:scaling}), where $K$ is the augmented-set size relative to the original data ($K{=}10$ is a $10\times$ expansion).
Holding the backbone (SpaMo~\cite{hwang2025spamo}) and training epochs fixed (\cref{sec:benchmark_aug_eval}), BLEU-4 increases monotonically with $K$ on both Phoenix14T and How2Sign~\cite{camgoz2018neural, duarte2021how2sign} ($14.37\!\rightarrow\!18.54$ and $10.23\!\rightarrow\!14.11$), indicating that larger augmentation budgets yield stronger robustness and motivating the scaling of SignNet-1M to million-level volume. Matched-GPU-hour comparisons are provided in the Appendix.

\subsection{Visual Quality Assessment}
\label{sec:exp_quality}

\textbf{Qualitative analysis.}
\cref{fig:augmentation_examples} visualizes representative outputs across the pipeline: novel-view renderings preserve hand shape and facial expression under pitch/yaw/zoom/dynamic-camera shifts; cross-identity reenactment maintains motion synchrony across signer appearances; background replacement produces realistic scene context; and post-rendering augmentations simulate geometric, photometric, and temporal degradations, while largely preserving linguistic content and motion coherence.

\textbf{Quantitative analysis.}
We assess visual quality with distribution-level FID-VID~\cite{heusel2017gans}/FVD~\cite{unterthiner2019fvd} and pairwise PSNR/SSIM~\cite{wang2004ssim} (full metrics in the supplementary material).
SignNet-Aug stays close to the originals: relative to Origin, FID-VID changes within $[-2.5\%, +3.0\%]$, FVD within $[-23.0\%, +15.4\%]$, and SSIM remains high ($\ge 0.81$), indicating strong distributional realism and structural consistency.

\textbf{Identity-swap fidelity.} To verify cross-identity reenactment does not degrade usability (Fig.~\ref{fig:augmentation_examples}e), we compare identity-swapped against non-identity-swapped clips (full numbers in the supplementary material): FID-VID shifts by only ${\sim}2\%$ and translation stays comparable (e.g., How2Sign BLEU-4 $10.11\!\rightarrow\!12.54$), confirming that identity-swapped videos preserve fine-grained articulation and temporal consistency.

\subsection{Human Evaluation}
\label{sec:exp_human}
To verify that augmentation preserves sign semantics and linguistic annotations, we conducted two IRB-approved studies with 14 hired native-ASL signers (details in the supplementary material). (i) On 27{,}000 source/synthesized ASL sentence pairs from How2Sign~\cite{duarte2021how2sign}, OpenASL~\cite{shi2022openasl}, and ASL50K~\cite{asl50k}, novel-view, background, and post-rendering augmentations reached $>99\%$ acceptance, and the hardest task, cross-identity reenactment, $94.8\%$. (ii) On a self-collected set of 175 ASL sentences and 500 isolated words, acceptance was $100\%$. With the downstream gains (\cref{tab:main_exp,tab:main_exp_slr}), this confirms semantic preservation.

%% file: sections_v0305/08_limitations_future_work.tex
\section{Limitations}
\label{sec:limitations_future}
(1)~Since GUAVA ~\cite{zhang2025guava} reconstructs avatars from a single image, rendering quality may degrade under large viewpoint changes, especially for fine hand details and loose clothing.
(2)~SignNet-1M includes three sign languages (DGS/ASL/CSL); results may not fully generalize to other languages.
(3)~Identity images for cross-reenactment are drawn solely from publicly released corpora with research consent; synthesized clips are metadata-tagged to prevent re-identification misuse.
(4)~To enable broader evaluation across model families beyond the methods benchmarked here, we will host a public \emph{SignNet-1M challenge} so the community can assess additional architectures on the unified protocol.

%% file: sections_v0305/09_conclusion.tex
\section{Conclusion}
\label{sec:conclusion}

We presented SignNet-1M, a large-scale multilingual augmented sign language dataset with ${\sim}$1M clips across ASL, DGS, and CSL.
By combining 3DGS-based novel-view rendering, diffusion-based scene/identity editing, and lightweight post-rendering augmentations, SignNet-1M creates controlled shifts in viewpoint, background, and signer appearance while preserving sign motion and annotations.
Our Orig / Zero-shot / Trained evaluation reveals substantial robustness gaps under distribution shifts and shows that SignNet-1M improves robustness for real-world deployment.
We release the dataset, the full augmentation pipeline (code and per-stage configs), and evaluation scripts to support reproducible research at \url{https://signnet.chatsign.ai/}.

%% file: sections_v0305/appendix.tex
\appendix
\section{Additional Implementation Details}

\subsection{Augmentation Hyperparameters}
\label{app:aug_hparams}

Table~\ref{tab:aug_hparams} summarizes the concrete settings used in our released augmentation pipeline.
We report the choices that materially affect the generated data distribution; unless otherwise stated, all other options follow the default configurations of the corresponding released codebases.

\begin{table}[t]
\centering
\footnotesize
\renewcommand{\arraystretch}{1.12}
\setlength{\tabcolsep}{3pt}
\begin{tabularx}{\linewidth}{@{}>{\raggedright\arraybackslash}p{2.45cm}>{\raggedright\arraybackslash}X@{}}
\toprule
\textbf{Component} & \textbf{Key settings} \\
\midrule

Background editing
& Five-stage background-editing pipeline at $848\times480$ resolution; representative preprocessing, IC-Light~\cite{zhang2025iclight}, and FlowPortal~\cite{gao2025flowportal} settings are summarized in \cref{app:bg_hparams}. Lighting severity is defined separately in \cref{app:lighting_definition}. \\

EHM-Tracker~\cite{zhang2025ehmtracker}
& Crop sizes (body/HD/head/hand) $=224/1024/512/512$; teaser input $=224$; FPS $=30$; CRF $=15$; perspective projection with $\tan(\mathrm{fov})=1/24$. \newline
  FLAME fitting, EHM refinement, optimization, and VPoser enabled; refinement steps $=1001$. \\

GUAVA~\cite{zhang2025guava}
& $\texttt{uvmap\_size}=512$; $\texttt{image\_size}=512$; $\texttt{feature\_img\_size}=518$; batch size $=6$; training iterations $=200$k; seed $=10$. \newline
  Adam with lr $=10^{-4}$ and $\beta=(0.0,0.99)$; linear learning-rate decay. \\

Novel-view
& Yaw/pitch/zoom/dynamic/mixed variants via GUAVA. Static-view ranges follow the main paper: yaw $\in[-30^{\circ},30^{\circ}]$, pitch $\in[-25^{\circ},25^{\circ}]$, and zoom $r\in[0.6,1.5]$; dynamic views sample two endpoints from the same static range and linearly interpolate over time. \\

Post-rendering
& Following the main paper, post-rendering uses a $\times 12$ mix with origin (20\%), geometric (24\%: crop, rotate, perspective), color (28\%: brightness, saturation, gamma), degradation (16\%: blur, noise, compression), and temporal (12\%: frame-rate drop, speed perturbation, low-fps sampling). \\

Cross-identity
& Source identity from the first frame of the source clip; motion/expression from the target sequence. \newline
  Identity/body-shape-related SMPL-X/FLAME~\cite{pavlakos2019smplx,li2017flame} parameters are copied from source to target; target-motion camera by default; full-video rendering at $30$ FPS. \\

\bottomrule
\end{tabularx}
\caption{Key hyperparameters of the augmentation pipeline.}
\label{tab:aug_hparams}
\end{table}

\subsection{Background Editing Settings}
\label{app:bg_hparams}

The representative background-editing settings referenced in Table~\ref{tab:aug_hparams} are detailed here.
Our background-editing pipeline consists of five stages:
(i) uniform frame resampling and control extraction,
(ii) foreground-mask estimation,
(iii) mask-aware second-pass preprocessing,
(iv) IC-Light~\cite{zhang2025iclight} reference generation, and
(v) FlowPortal~\cite{gao2025flowportal} video editing.

During preprocessing, frames are sampled uniformly at the target frame rate, and clips shorter than the target length are discarded.
We use batch size $7$ and \texttt{edge\_mode=combined}, where the control signal is formed as
$0.1\times$ Canny $+0.1\times$ HED $+0.8\times$ depth; the Canny thresholds are $(50,100)$.
Foreground masks are estimated with BiRefNet~\cite{zheng2024birefnet}.

For illumination-aware reference generation, IC-Light~\cite{zhang2025iclight} uses a Realistic Vision v5.1 base model with a DPMSolverMultistep scheduler, guidance scale $2.0$, high-resolution scale $1.5$, and high-resolution denoising strength $0.5$.
We use $25$ denoising steps in the batch pipeline and $50$ in the standalone script.

For temporally consistent video editing, FlowPortal uses the released Wan2.1 configuration.
In batch mode, we use $25$ sampling steps ($50$ in the standalone script), source/target guidance scales $6.0/6.0$, edit amplifier $1.0$, partial editing, and source/transfer blur strengths $0.5/0.4$.
Acceleration uses \texttt{cache\_times}=10, TeaCache threshold $0.10$, and \texttt{cfg\_skip\_ratio}=0.15.

\subsection{Training and Evaluation Settings}

Table~\ref{tab:train_eval_settings} summarizes the main training and evaluation settings of the two evaluated backbones.

\begin{table}[t]
\centering
\footnotesize
\renewcommand{\arraystretch}{1.1}
\setlength{\tabcolsep}{3pt}
\begin{tabularx}{\linewidth}{@{}>{\raggedright\arraybackslash}p{2.15cm}>{\raggedright\arraybackslash}X>{\raggedright\arraybackslash}X@{}}
\toprule
\textbf{Setting} & \textbf{SpaMo} & \textbf{UniSign} \\
\midrule
Input / features
& CLIP ViT-L/14~\cite{radford2021clip} spatial + VideoMAE-L~\cite{tong2022videomae} motion features; multi-scale spatial extraction and sliding-window temporal extraction.
& RTMPose WholeBody~\cite{jiang2023rtmpose} (133 keypoints); body/left-hand/right-hand/face streams; max $256$ frames; body-scale normalization; confidence threshold $0.3$. \\

Backbone
& Flan-T5-XL~\cite{chung2024flan} with LoRA~\cite{hu2022lora} on query/value only ($r=16$, $\alpha=32$, dropout $=0.1$).
& ST-GCN encoder + mT5-Base~\cite{xue2021mt5} decoder; 64-d keypoint projection; fused to the 768-d mT5 hidden space. \\

Core settings
& Input size $2048$; hidden size $768$; joint fusion; max frames $512$; max text $64$; frame sample rate $1$; 3-shot in-context prompting; cross-modal alignment + combined loss.
& Label smoothing $0.2$; pose-only translation by default. \\

Optimization
& AdamW, lr $=6\times 10^{-4}$, wd $=0.01$, $\beta=(0.9,0.98)$, $\epsilon=10^{-8}$; cosine, 10\% warmup.
& AdamW, lr $=10^{-3}$, wd $=10^{-4}$, $\epsilon=10^{-9}$; cosine; DeepSpeed ZeRO-2~\cite{rasley2020deepspeed,rajbhandari2020zero}, bf16, grad clip $1.0$. \\

Training setup
& Batch $8 \times 2$ accumulation (effective $16$); bf16; grad clip $1.0$; up to $500$ epochs; early stopping patience $50$ on val BLEU-4.
& CSL-Daily~\cite{zhou2021signbt}: batch $128$, accum $1$; How2Sign~\cite{duarte2021how2sign}: batch $64$, accum $2$; effective batch $128$; seed $42$. \\

Evaluation
& Beam $5$, max length $64$, \texttt{top\_p}$=0.9$; sacreBLEU BLEU-4~\cite{papineni2002bleu}.
& Beam $4$, \texttt{max\_new\_tokens}$=100$; BLEU-1/2/3/4~\cite{papineni2002bleu}, ROUGE~\cite{lin2004rouge} (SLT), and WER\% (SLR). \\
\bottomrule
\end{tabularx}

\caption{Main training and evaluation settings of SpaMo~\cite{hwang2025spamo} and UniSign~\cite{li2025unisign}.}
\label{tab:train_eval_settings}
\end{table}

\subsection{Hardware Infrastructure}

Table~\ref{tab:hardware_infra} summarizes the hardware and software environment used for training.
All experiments were run on an NVIDIA DGX Spark cluster managed by Slurm.
The cluster contains 30 compute nodes in the default \texttt{spark} partition.
Each node provides a single NVIDIA GB10 GPU with approximately 120\,GB memory, 20 ARM CPU cores, and about 120\,GB system memory.
All training runs use single-GPU jobs, with mixed-precision training enabled by default.

\begin{table}[t]
\centering
\footnotesize
\renewcommand{\arraystretch}{1.1}
\setlength{\tabcolsep}{3pt}
\begin{tabularx}{\linewidth}{@{}>{\raggedright\arraybackslash}p{2.6cm}>{\raggedright\arraybackslash}X@{}}
\toprule
\textbf{Item} & \textbf{Specification} \\
\midrule
Cluster
& NVIDIA DGX Spark cluster; 30 compute nodes. \\

Per-node accelerator
& 1$\times$ NVIDIA GB10 (Blackwell), 120GB GPU memory, CUDA 13.0, NVIDIA driver 580.95.05. \\

Per-node host
& 20 ARM CPU cores (\texttt{aarch64}), 120\,GB system memory, Ubuntu 24.04.4 LTS. \\

Shared storage
& CIFS/SMB-backed storage with 2.2\,PB total capacity. \\

Training software
& Python (conda), PyTorch 2.10.0+cu128, DeepSpeed 0.18.6, Transformers 4.40.0, CUDA Toolkit 13.0, timm 0.8.10.dev0. \\

\bottomrule
\end{tabularx}

\caption{Hardware and software environment used for dataset generation, training and evaluation.}
\label{tab:hardware_infra}
\end{table}

\subsection{Computational Overhead}
\label{app:compute}
Table~\ref{tab:compute} reports the average per-clip cost of each augmentation stage (single NVIDIA DGX B200 / DGX Spark). EHM-Tracker fitting is a shared preprocessing step amortized across all downstream augmentations. Generating the full SignNet-1M corpus costs approximately 12K GPU-hours.
\begin{table}[t]
\centering
\footnotesize
\renewcommand{\arraystretch}{1.1}
\setlength{\tabcolsep}{4pt}
\begin{tabularx}{\linewidth}{@{}>{\raggedright\arraybackslash}Xll@{}}
\toprule
\textbf{Stage} & \textbf{Avg.\ cost / clip} & \textbf{Notes} \\
\midrule
EHM-Tracker fitting       & ${\sim}4$ min & preprocess (shared) \\
Novel-view rendering      & ${\sim}3$ s   & reuses 3DGS assets \\
Scene editing (background) & ${\sim}3$ min & background replacement \\
Identity reenactment      & ${\sim}5$ s   & signer swap \\
Post-rendering aug.        & ${\sim}2$ s   & perturbations \\
\midrule
\multicolumn{3}{@{}l}{Full SignNet-1M generation: ${\sim}$12K GPU-hours.} \\
\bottomrule
\end{tabularx}
\caption{Per-clip computational overhead of each augmentation stage.}
\label{tab:compute}
\end{table}

\FloatBarrier
\section{Additional Quantitative Results}
This section complements the main paper with full metric breakdowns and a matched-compute view.

\subsection{Full Sign Language Translation Results}
\label{app:full_slt_results}

Table~\ref{tab:full_slt_results} reports full SLT metrics for SpaMo~\cite{hwang2025spamo} under the \textit{Orig.}, \textit{Zero-shot}, and \textit{Trained} protocols.
We report results on Phoenix14T~\cite{camgoz2018neural}, How2Sign~\cite{duarte2021how2sign}, OpenASL~\cite{shi2022openasl}, and CSL-Daily~\cite{zhou2021signbt}.
Beyond BLEU-4, the same gap-and-gain pattern also appears in BLEU-1/2/3 and ROUGE-L~\cite{lin2004rouge}: zero-shot evaluation consistently underperforms the original setting, while training on SignNet-1M recovers a substantial portion of the loss.
Across the four sub-datasets, the zero-shot gap ranges from 3.86 to 14.68 BLEU-4 points and from 8.10 to 27.92 RG points; training on SignNet-1M then recovers 4.11--11.17 BLEU-4 points and 9.09--26.92 RG points.
These full-metric trends further support the main-paper conclusion that SignNet-1M is necessary for robustness under realistic distribution shifts, rather than only improving a single headline metric.

\begin{table*}[t]
    \centering
    \caption{Full SLT results on the test split for SpaMo~\cite{hwang2025spamo}. For conciseness, the current supplementary version reports SpaMo only. `B1--B4' denote BLEU-1/2/3/4, and `RG' denotes ROUGE-L~\cite{lin2004rouge} F1.}
    \label{tab:full_slt_results}
    \scriptsize
    \setlength{\tabcolsep}{4pt}
    \renewcommand{\arraystretch}{1.05}
    \begin{tabular}{@{}lllccccc@{}}
        \toprule
        Sub-dataset & Method & Setting & B1 & B2 & B3 & B4 & RG \\
        \midrule
        \multirow{3}{*}{Phoenix14T~\cite{camgoz2018neural}}
          & \multirow{3}{*}{SpaMo~\cite{hwang2025spamo}} & Orig.      & 48.12 & 35.19 & 27.42 & 22.49 & 44.19 \\
          &                        & Zero-shot  & 18.38 & 13.09 & 9.31 & 7.81 & 16.27 \\
          &                        & Trained    & 46.48 & 30.60 & 26.94 & 18.98 & 43.19 \\
        \midrule
        \multirow{3}{*}{How2Sign~\cite{duarte2021how2sign}}
          & \multirow{3}{*}{SpaMo~\cite{hwang2025spamo}}   & Orig.      & 33.41 & 20.28 & 13.96 & 10.11 & 30.56 \\
          &                          & Zero-shot  & 19.35 & 15.37 & 8.79 & 6.25 & 18.00 \\
          &                          & Trained    & 33.63 & 19.60 & 13.25 & 10.36 & 31.02 \\
        \midrule
        \multirow{3}{*}{OpenASL~\cite{shi2022openasl}}
          & \multirow{3}{*}{SpaMo~\cite{hwang2025spamo}}   & Orig.      & 36.03 & 22.93 & 15.74 & 10.32 & 32.21 \\
          &                          & Zero-shot  & 7.49 & 5.13 & 1.49 & 0.34 & 5.16 \\
          &                          & Trained    & 34.43 & 20.58 & 15.74 & 9.71 & 30.48 \\
        \midrule
        \multirow{3}{*}{CSL-Daily~\cite{zhou2021signbt}}
          & \multirow{3}{*}{SpaMo~\cite{hwang2025spamo}}   & Orig.      & 48.90 & 36.90 & 26.78 & 20.55 & 47.46 \\
          &                          & Zero-shot  & 40.08 & 28.82 & 20.24 & 12.70 & 39.36 \\
          &                          & Trained    & 50.24 & 37.20 & 28.95 & 21.32 & 48.45 \\
        \bottomrule
    \end{tabular}
\end{table*}

\subsection{Full Sign Language Recognition Results}
\label{app:full_slr_results}

Table~\ref{tab:full_slr_results} reports full SLR metrics for Online-CSLR~\cite{zuo2024towards} on Phoenix14T~\cite{camgoz2018neural} and UniSign~\cite{li2025unisign} on CSL-Daily~\cite{zhou2021signbt}, including WER\% and its deletion, insertion, and substitution components.
The same conclusion also holds for recognition: the zero-shot setting increases WER\% and all three error components, while training on SignNet-1M reduces these errors substantially.
Numerically, zero-shot raises WER\% by 29.24\% and 28.70\% on Phoenix14T and CSL-Daily, respectively; deletion, insertion, and substitution rates increase by 9.08\%--10.92\%, 1.83\%--3.18\%, and 16.44\%--16.49\%. Training then reduces WER\% by 23.26\%--26.16\% relative to zero-shot.
In particular, the decomposition into deletion, insertion, and substitution rates shows that the recognition gap is not confined to a single failure mode, again highlighting the necessity of the proposed dataset.

\begin{table}[t]
    \centering
    \caption{Full SLR results on the test split for Online-CSLR~\cite{zuo2024towards} and UniSign~\cite{li2025unisign}. In addition to WER\%, we report deletion, insertion, and substitution rates. All values are percentages.}
    \label{tab:full_slr_results}
    \scriptsize
    \setlength{\tabcolsep}{4pt}
    \renewcommand{\arraystretch}{1.05}
    \begin{tabular}{@{}lllcccc@{}}
        \toprule
        Sub-dataset & Method & Setting & WER\% & Del.\% & Ins.\% & Sub.\% \\
        \midrule
        \multirow{3}{*}{Phoenix14T~\cite{camgoz2018neural}}
          & \multirow{3}{*}{Online-CSLR~\cite{zuo2024towards}} & Orig.      & 22.21\% & 6.09\% & 1.99\% & 14.13\% \\
          &                               & Zero-shot  & 51.45\% & 17.01\% & 3.82\% & 30.62\% \\
          &                               & Trained    & 28.19\% & 7.96\% & 2.67\% & 17.56\% \\
        \midrule
        \multirow{3}{*}{CSL-Daily~\cite{zhou2021signbt}}
          & \multirow{3}{*}{UniSign~\cite{li2025unisign}}     & Orig.      & 28.20\% & 9.27\% & 2.84\% & 16.09\% \\
          &                              & Zero-shot  & 56.90\% & 18.35\% & 6.02\% & 32.53\% \\
          &                              & Trained    & 30.74\% & 8.31\% & 4.58\% & 17.85\% \\
        \bottomrule
    \end{tabular}
\end{table}

\subsection{Generalization Across Model Families}
\label{app:model_family}
Table~\ref{tab:model_family} summarizes the improvement of SignNet-1M-trained over original-dataset-trained models across three typical model-family specifications used in the sign-language community (RGB gloss-free, RGB+Pose gloss-based SLT, and pose-based recognition). The consistent gains indicate that the benefit of SignNet-1M is not tailored to a single architecture.

\begin{table}[t]
    \centering
    \small
    \setlength{\tabcolsep}{4pt}
    \caption{Generalization across model families. $\Delta$ is the improvement of SignNet-1M-trained over original-dataset-trained models (BLEU-4 for SLT methods, WER\% for the Online-CSLR recognizer). SOTA methods span three typical model-family specifications used in the sign-language community.}
    \label{tab:model_family}
    \begin{tabular}{llccc}
        \toprule
        Method & Modality & Supervision & Task & $\Delta$ \\
        \midrule
        SpaMo~\cite{hwang2025spamo}      & RGB      & Gloss-free  & SLT & +11.17 \\
        FPT4SLT~\cite{wong2024sign2gpt}  & RGB      & Gloss-free  & SLT & +11.47 \\
        GASLT~\cite{yin2023gaslt}        & RGB      & Gloss-free  & SLT & +13.69 \\
        UniSign~\cite{li2025unisign}     & RGB+Pose & Gloss-based & SLT & +14.71 \\
        Online-CSLR~\cite{zuo2024towards}& RGB+Pose & Gloss-based & SLR & +23.26\% \\
        \bottomrule
    \end{tabular}
\end{table}

\subsection{Visual Quality Metrics}
\label{app:visual_quality}
Table~\ref{tab:quality} compares the distributional and pairwise visual quality of original versus SignNet-augmented videos across all five sub-datasets. Relative to Origin, FID-VID changes within $[-2.5\%, +3.0\%]$ and FVD within $[-23.0\%, +15.4\%]$, while SSIM stays high ($\ge 0.81$), indicating that augmentation preserves distributional realism and structural consistency.

\begin{table}[t]
    \centering
    \caption{Visual quality comparison between original and SignNet-augmented videos. Lower FID-VID/FVD and higher SSIM/PSNR indicate better quality.}
    \label{tab:quality}
    \small
    \setlength{\tabcolsep}{5pt}
    \begin{tabular}{lcccccc}
        \toprule
        \multirow{2}{*}{Sub-Dataset}
        & \multicolumn{2}{c}{FID-VID$\downarrow$}
        & \multicolumn{2}{c}{FVD$\downarrow$}
        & \multirow{2}{*}{SSIM$\uparrow$}
        & \multirow{2}{*}{PSNR$\uparrow$} \\
        \cmidrule(lr){2-3} \cmidrule(lr){4-5}
        & Origin & SignNet-Aug & Origin & SignNet-Aug & & \\
        \midrule
        Phoenix14T~\cite{camgoz2018neural}      & 49.96 & \textbf{49.22} & 141.57 & \textbf{119.82} & 0.92 & 25.30 \\
        How2Sign~\cite{duarte2021how2sign}   & \textbf{47.26} & 48.68 & \textbf{118.69} & 123.84 & 0.91 & 27.98 \\
        ASL50K~\cite{asl50k}                               & \textbf{43.41} & 44.06 & \textbf{119.15} & 137.52 & 0.83 & 21.83 \\
        OpenASL~\cite{shi2022openasl}        & 49.52 & \textbf{48.29} & 157.79 & \textbf{121.44} & 0.81 & 23.00 \\
        CSL-Daily~\cite{zhou2021signbt}      & 40.39 & \textbf{39.80} & \textbf{124.45} & 136.34 & 0.88 & 23.24 \\
        \bottomrule
    \end{tabular}
\end{table}

\subsection{Identity-Swap Fidelity}
\label{app:identity_quality}
Table~\ref{tab:identity_quality} compares identity-swapped against non-identity-swapped clips. FID-VID shifts by only ${\sim}2\%$ (Phoenix14T $50.99\!\rightarrow\!49.67$, $-0.6\%$; How2Sign $47.26\!\rightarrow\!48.05$, $+1.7\%$; CSL-Daily $40.39\!\rightarrow\!39.61$, $-1.9\%$), and translation quality stays comparable (Phoenix14T BLEU-4 $22.04$ with vs.\ $22.49$ without identity editing, $-2.0\%$; How2Sign improves $10.11\!\rightarrow\!12.54$, $+24.0\%$), confirming that identity-swapped videos preserve fine-grained articulation and temporal consistency.

\begin{table}[t]
    \centering
    \small
    \setlength{\tabcolsep}{5pt}
    \caption{Identity-swap fidelity. FID-VID$\downarrow$ for identity-swapped vs.\ non-identity-swapped clips, and BLEU-4 of models trained with vs.\ without identity editing. Identity swapping changes FID-VID by only ${\sim}2\%$ and keeps translation quality comparable.}
    \label{tab:identity_quality}
    \begin{tabular}{lcccc}
        \toprule
        & \multicolumn{2}{c}{FID-VID$\downarrow$} & \multicolumn{2}{c}{BLEU-4$\uparrow$} \\
        \cmidrule(lr){2-3} \cmidrule(lr){4-5}
        Sub-dataset & w/o ID swap & w/ ID swap & w/o ID edit & w/ ID edit \\
        \midrule
        Phoenix14T~\cite{camgoz2018neural} & 50.99 & 49.67 & 22.49 & 22.04 \\
        How2Sign~\cite{duarte2021how2sign} & 47.26 & 48.05 & 10.11 & 12.54 \\
        CSL-Daily~\cite{zhou2021signbt}    & 40.39 & 39.61 & --    & --    \\
        \bottomrule
    \end{tabular}
\end{table}

\subsection{Human Evaluation}
\label{app:human_eval}
Table~\ref{tab:human_eval} details the two IRB-approved human studies conducted with 14 hired native-ASL signers. (i) On 27{,}000 ASL source/synthesized sentence pairs from How2Sign~\cite{duarte2021how2sign}, OpenASL~\cite{shi2022openasl}, and ASL50K~\cite{asl50k}, novel-view, background, and post-rendering augmentations achieved $>99\%$ acceptance, and the hardest task, cross-identity reenactment, reached $94.8\%$. (ii) On a self-collected high-quality set of 175 ASL sentences and 500 isolated words, acceptance was $100\%$.

\begin{table}[t]
    \centering
    \small
    \setlength{\tabcolsep}{5pt}
    \caption{Human evaluation by 14 native-ASL signers. Acceptance = fraction of synthesized clips judged to preserve the source sign and its linguistic content.}
    \label{tab:human_eval}
    \begin{tabular}{lcc}
        \toprule
        Study / augmentation & \#Samples & Acceptance \\
        \midrule
        Novel-view / background / post-rendering & \multirow{2}{*}{27{,}000 pairs} & $>99\%$ \\
        Cross-identity reenactment               &                                 & $94.8\%$ \\
        \midrule
        Self-collected (175 sentences, 500 words) & 675 & $100\%$ \\
        \bottomrule
    \end{tabular}
\end{table}

\subsection{Matched-Compute Comparison}
\label{app:matched_compute}

\begin{figure}[t]
    \centering
    \includegraphics[width=\linewidth]{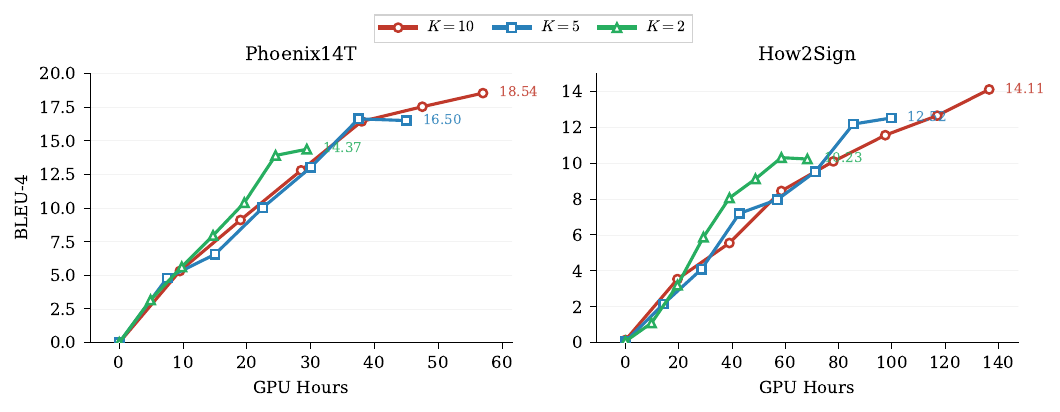}
    \caption{Matched-compute comparison of SpaMo~\cite{hwang2025spamo} under increasing augmentation scale $K$. While larger $K$ achieves the highest final BLEU-4, the curves largely overlap under comparable GPU-hour budgets, indicating that the benefit of larger-scale augmentation comes from both additional optimization and increased data diversity.}
    \label{fig:matched_compute}
\end{figure}

Figure~\ref{fig:matched_compute} adds a matched-compute view of BLEU-4 versus GPU hours under different augmentation scales $K$, complementing the epoch-based scaling curves in the main paper.
Viewed by epoch, larger $K$ consistently reaches the highest BLEU-4.
Under matched GPU-hour budgets, however, the curves are much closer; on Phoenix14T, $K{=}2$ is even slightly better at comparable compute.
$K{=}2$ and $K{=}5$ saturate early, whereas $K{=}10$ continues improving and achieves the best final performance.
This suggests that the gains from SignNet-1M are not explained solely by extra optimization, but also by the added diversity of larger-scale augmentation, at the cost of higher GPU-hour consumption.

\section{Severity Experiments}
\subsection{Exact Definition of Lighting Severity}
\label{app:lighting_definition}

For the background-editing benchmark, lighting severity is not defined by manual semantic labels (e.g., ``daytime'' or ``night'').
Instead, we assign each edited clip a continuous photometric-shift score that measures how strongly the edited video deviates from its original source video in signer-region appearance.

Specifically, each edited video in \texttt{aug\_background} is paired with its corresponding original video.
For every frame pair, we convert both frames to the Lab color space and compute frame-level masked mean statistics over the signer region defined by the foreground mask.
Let $\mu_L$, $\mu_a$, and $\mu_b$ denote the masked mean Lab channels of a frame.
We define the frame-level luminance difference as
\begin{equation}
\Delta_L = \frac{\left| \mu_L^{\text{edit}} - \mu_L^{\text{orig}} \right|}{255},
\end{equation}
and the frame-level chromatic difference as
\begin{equation}
\Delta_{ab} =
\sqrt{
\left(\mu_a^{\text{edit}} - \mu_a^{\text{orig}}\right)^2 +
\left(\mu_b^{\text{edit}} - \mu_b^{\text{orig}}\right)^2 }.
\end{equation}

These differences are averaged over all paired frames in the clip to obtain a video-level luminance shift $\overline{\Delta}_L$ and chromatic shift $\overline{\Delta}_{ab}$.
We then define the raw lighting-shift score as
\begin{equation}
s_{\text{raw}} = 0.7\,\overline{\Delta}_L + 0.3\,\frac{\overline{\Delta}_{ab}}{P_{95}^{\text{chroma}}},
\end{equation}
where $P_{95}^{\text{chroma}}$ is the 95th percentile of chromatic shifts computed over the dataset, used to normalize the chromatic term.
Finally, the score is normalized to $[0,1]$ by
\begin{equation}
s_{\text{light}} = \mathrm{clip}\!\left( \frac{s_{\text{raw}}}{P_{95}^{\text{raw}}},\, 0,\, 1 \right),
\end{equation}
where $P_{95}^{\text{raw}}$ is the 95th percentile of raw scores over the dataset.

For evaluation, we discretize $s_{\text{light}}$ into ten equal-width bins over $[0,1]$.
Therefore, levels L1--L10 correspond to progressively larger signer-region photometric shifts, rather than predefined semantic or physical lighting categories.
